\documentclass[11pt]{article}
\usepackage{sigsam, amsmath}
\usepackage{graphicx}
\usepackage{float}

\issue{TBA}
\articlehead{}
\titlehead{Symbolics.jl}
\authorhead{Gowda \textit{et al.}}
\begin{document}

\title{High-performance symbolic-numerics via multiple dispatch}

\author{Shashi Gowda$^{1}$, Yingbo Ma$^{2}$, Alessandro Cheli$^{3}$, Maja Gw\'{o}\'{z}d\'{z}$^{2}$\\  Viral B. Shah$^{2}$, Alan Edelman$^{1}$, Christopher Rackauckas$^{1,2}$ \\
$^{1}$ Massachusetts Institute of Technology \\
$^{2}$ Julia Computing \\
$^{3}$ University of Pisa, Pisa, Italy}

\date{}

\maketitle

\vspace{-30px}
\begin{abstract}
As mathematical computing becomes more democratized in high-level languages, high-performance symbolic-numeric systems are necessary for domain scientists and engineers to get the best performance out of their machine without deep knowledge of code optimization. Naturally, users need different term types either to have different algebraic properties for them, or to use efficient data structures. To this end, we developed Symbolics.jl, an extendable symbolic system which uses dynamic multiple dispatch to change behavior depending on the domain needs. In this work we detail an underlying abstract term interface which allows for speed without sacrificing generality. We show that by formalizing a generic API on actions independent of implementation, we can retroactively add optimized data structures to our system without changing the pre-existing term rewriters. We showcase how this can be used to optimize term construction and give a 113x acceleration on general symbolic transformations. Further, we show that such a generic API allows for complementary term-rewriting implementations. Exploiting this feature, we demonstrate the ability to swap between classical term-rewriting simplifiers and e-graph-based term-rewriting simplifiers. We illustrate how this symbolic system improves numerical computing tasks by showcasing an e-graph ruleset which minimizes the number of CPU cycles during expression evaluation, and demonstrate how it simplifies a real-world reaction-network simulation to halve the runtime. Additionally, we show a reaction-diffusion partial differential equation solver which is able to be automatically converted into symbolic expressions via multiple dispatch tracing, which is subsequently accelerated and parallelized to give a 157x simulation speedup. Together, this presents Symbolics.jl as a next-generation symbolic-numeric computing environment geared towards modeling and simulation.
\end{abstract}

\section{Introduction}

Writing fast code is not the domain of humans, instead it is the domain of symbolic engines. Many core numerical routines, such as the FFTW suite for Fast Fourier Transformations, make use of symbolic calculations to vastly outperform purely handwritten methods \cite{frigo2005design}. Additionally, techniques like Herbie show how to improve numerical stability in a semi-automated fashion via symbolic rule applications \cite{panchekha2015automatically}. However, these techniques currently require sophisticated programmers to build domain-specific and scalable code optimizers. In this work we explore the question --- could these mathematical code optimizations be generally applied in an automated fashion to a whole high-performance programming language? 

To meet this goal, we developed Symbolics.jl, a high-performance symbolic-numeric computing library with a type-dispatch system to give the generality necessary to target the full Julia programming language. Section \ref{sec:generality} shows how Symbolics.jl's simple and non-intrusive term interface solves the Expression Problem in symbolic computing and allows for extending the core functionality without inhibiting performance. Section \ref{sec:rewriting} showcases how this generic term interface allows for multiple complementary term rewriting systems to coexist, which allows for mixing modern e-graph code optimizers with traditional simplifiers. Section \ref{sec:codegen} discusses the code generation capabilities of Symbolics.jl. Given that our language is the JIT-compiled Julia language, the emitted code is compiled and executed in a fast runtime. This allows for both fast debug cycles and for using Julia's well-optimized numerical ecosystem with built-in parallelism in the generated code. Section \ref{sec:results} details instances where scientific codes can be automatically optimized using this mechanism.


\section{Generality without sacrificing performance}

\label{sec:generality}

Many programming systems suffer from the Expression Problem \cite{FF}, which is the inability to specialize functions on types not defined when a module is compiled. Dynamic multiple dispatch solves this by allowing functions to be specialized for types that are loaded at runtime effectively allowing developers to decouple data representations from actions on them \cite{bezanson2017julia}. Given that the Julia ecosystem significantly leverages multiple dispatch, we used it to develop a similarly generic yet high-performance system for symbolic computing. However, we do note that the ideas in this paper can also be implemented in any language with a sufficiently powerful generic programming capabilities, such as Haskell or Common Lisp.

The core of a symbolic library is its representation of mathematical terms. The choice of term representation can impose restrictions on generality. For example, the simplest term representation can be found in Lisp-like systems: terms are quoted expressions, which, in turn, are simply lists of lists or atoms. While this is an elegant and simple solution, it has a few shortcomings, most importantly:

\begin{enumerate}
\vspace{-8pt}
    \item The user can, at best, define one overloading of $+$ and $*$ for all expressions, but expressions can be of different types. This is a symptom of the Expression Problem.
\vspace{-8pt}
    \item The size of a term grows linearly with the number of operations unless terms are simplified during construction, which can be expensive using this data structure.
\end{enumerate}
\vspace{-10pt}

A solution to the first shortcoming is to use a parameterized \texttt{Term\{T\}(f, args)} struct to encode terms. Here, \texttt{T} is a type parameter, the symbolic type of the expression. We define operators, such as \texttt{+} and \texttt{*}, on the \texttt{Term\{<:Number\}} (numbers), and leave them open to be extended for non-number \texttt{Term}s. This allows the users to overload subsequent operations using Julia's multiple dispatch mechanism, and thus specialize the symbolic behavior in a manner that is dependent on the type of object being acted on. The solution is further refined later in this section, after a motivating example.

To solve the second shortcoming, Symbolics.jl employs a number of constructor-based simplification mechanisms. Multiplication and addition of numbers are the most common operations, yet simplifying commutativity and associativity in a rule-based system takes a long time. Instead, we use an idea from SymEngine\footnote{\url{https://symengine.org/design/design.html}} which is to formulate a canonical form in terms of \texttt{Add} and \texttt{Mul} types, which simplify expressions upon construction. \texttt{Add} represents a linear combination of terms: it stores the numeric coefficients and their corresponding terms in a dictionary as values and keys, respectively. \texttt{Mul} stores a product of factor terms: it stores bases and the corresponding exponents in a dictionary as keys and values, respectively. This allows us to use $O(1)$ dictionary lookups to simplify repeated addition and multiplication. In the best case, they take up $O(1)$ space, while \texttt{Term} would take $O(n)$ space for $n$ operations.

Having introduced these types, one may think we have lost the generality provided by a common term type on which all symbolic manipulation code can rely. However, this generality is regained by defining the following set of generic functions, which form an interface that a term should satisfy:

\begin{enumerate}
    \vspace{-8pt}
    \item \texttt{istree(x)} -- returns \texttt{true} if \texttt{x} is a term object.
    \vspace{-8pt}
    \item \texttt{operation(x)} -- returns the head (the function) of the term.
    \vspace{-8pt}

    \item \texttt{arguments(x)} -- returns a vector of arguments of the term.
    \vspace{-8pt}
    \item \texttt{symtype(x)} -- returns the symbolic type of the term.
    \item \texttt{similarterm(x, f, args[, symtype])} -- constructs a term similar in type to \texttt{x} with \texttt{f} as head and \vspace{-8pt}
    \texttt{args} as arguments.
\end{enumerate}

For example, for a term \texttt{t} of type \texttt{Add}, \texttt{istree(t)} returns true, and \texttt{operation(t)} returns the \texttt{+} generic function. \texttt{arguments(t)} returns all the terms of the linear combination multiplied by the coefficient sorted according to an arbitrarily chosen total ordering. Finally, \texttt{symtype} represents the appropriate type, which is set when the \texttt{t} is created.

Term manipulation code can use \texttt{operation} and \texttt{arguments} after checking to see if \texttt{istree} returns \texttt{true} on an object. Such code can also use \texttt{similarterm} function to create a term. In fact, \texttt{Add} and \texttt{Mul} were added to our system retroactively based on the above interface, and the term manipulation codes continued to work as they did. The next section shows an example of our rule-based rewriting system. The system uses the above interface to access and construct terms, and thus does not require knowledge about the internal storage of terms.

\section{Term rewriting}

\label{sec:rewriting}

The abstract term interface allows for multiple rewrite mechanisms to co-exist simultaneously as part of the same symbolic platform. Here we describe two core rewriting systems that work on top of the term abstraction: a classical term rewriter which composes deterministic functions that turn an expression into a different one and a novel e-graph-based expression rewriting system which can apply bi-directional rules and optimize the rewritten expressions to minimize a cost function to achieve specific goals.

\subsection{Classical rewriting}

Classical rewriting is useful for traversing the expression tree in a specific way. In this regime, a rewriter is simply a function which takes an expression and returns a modified expression. Symbolics.jl provides a macro-based syntax that creates pattern-matching rewriters which act similarly to those described by Sussman and Hanson \cite{sdf}. To illustrate how pattern matching in performed, we demonstrate the macro on the double-angle formula. 
\vspace{-5pt}
\begin{verbatim}
r = @rule sin(2(~x)) => 2sin(~x)*cos(~x)
\end{verbatim}
\vspace{-5pt}
Here we use the \texttt{@rule} macro to create a rewriter to apply the canonical double-angle formula rule. \texttt{\textasciitilde x} on the left-hand side of the rule is called a ``slot'' and captures any object in the appropriate position in the expression tree. When \texttt{\textasciitilde x}\ appears on the right-hand side of the rule, the matched object is used in its place to perform the rewrite.

\vspace{-5pt}
\begin{verbatim}
@syms a::Real
r(2(deg2rad(a))) # => 2cos(deg2rad(a))*sin(deg2rad(a))
r(sin(3a)) # => 3a
\end{verbatim}
\vspace{-5pt}

Symbolics.jl contains a rewriter-combinator library for composing multiple rules. Briefly, \texttt{Chain([r1,r2])} combines two rewriters into one that applies \texttt{r1} and then \texttt{r2}. \texttt{IfElse(cond, r1, r2)} is a conditional rewriter. \texttt{Prewalk(r)} and \texttt{Postwalk(r)} take a rewriter \texttt{r} and produce a rewriter that traverses the tree and applies \texttt{r} to each tree node in pre-order or post-order, respectively. These walkers can also be run with multithreading to rewrite subtrees in parallel. Finally, \texttt{Fixpoint(r)} is a rewriter that applies \texttt{r} until there are no changes. The implementation of simplification on terms of \texttt{Number} subtypes uses 66 rules written in the rule-rewriting language, which are strung together by means of rewriter combinators. \label{sec:simplify}

\subsection{E-graph-based rewriting}
\begin{figure}[H]
    \centering
    \includegraphics[width=0.6\textwidth]{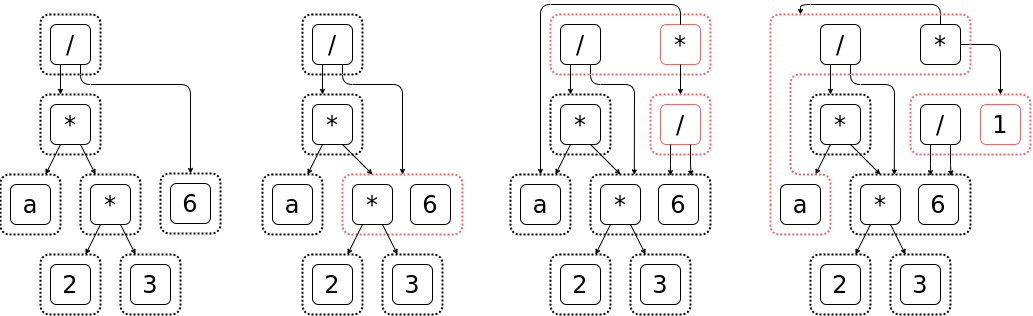}
    \caption{Equality saturation constructs the e-graph from a set of rules applied to an input expression. The four depicted e-graphs represent the process of equality saturation for the equivalent ways to write $a * (2 * 3) / 6$. The dashed boxes represent equivalence classes, and regular boxes represent e-nodes.}
    \label{fig:egraph}
\end{figure}

Classical rewriting falls short when the user's goal is to minimize a certain cost function and the system of rewrite rules comprises many rules interacting with one other. It is not straightforward to transform an axiomatic formal system of equational rules into a Noetherian (terminating) term-rewriting system, which is known to require a lot of user reasoning \cite{dershowitz1993taste}. To circumvent these issues, there has been newfound excitement in using equality saturation-based rewriting engines for such requirements. This novel technique allows users to define efficient term-rewriting systems with equational rules without having to worry about termination or the ordering of rules, resulting in algebraically compositional rewrite systems and efficient algorithms for minimizing cost functions on chains of expression rewrites. The Metatheory.jl \cite{Cheli2021} Julia package provides a generic rewriting backend relying on the \textit{equality saturation} algorithm  and data structures called \textit{e-graphs} (equality graphs) \cite{egg}. The intuitive mechanism behind \emph{e-graph rewriting} is explained briefly in Figure \ref{fig:egraph}. For more details, we refer the reader to \cite{Cheli2021} and \cite{egg}.

\section{High-performance code generation}

\label{sec:codegen}

Using Symbolics.jl, one can generate Julia code from symbolic expressions at run time, just-in-time compile them, and then execute them in the same session. This type of metaprogramming is much more convenient for mathematical code than macro-based metaprogramming. We expose a generic \texttt{toexpr} that turns expressions into executable Julia code. For convenience in generating sophisticated code, we have a library of term types that represent assignments, let-blocks, functions, array construction, and parallel map-reduce.

\section{Applications and results}

\label{sec:results}
\hspace{2em}\textbf{Overall speedup.} When tracing into the mass matrix computation of a rigid body dynamics system with 7 degrees of freedom, Symbolics.jl takes 0.0073 seconds, while SymPy takes 17.3 seconds. That is 2370x evaluated in a real-world application. Reproducible code can be found in the repository\footnote{\url{https://github.com/JuliaSymbolics/SymbolicUtils.jl/pull/254}}.

\textbf{Speedup from \texttt{Add} and \texttt{Mul}.} In a synthetic benchmark which generates a random expression of 1400 terms using $+$ and $*$, we get a speed up of 113$\times$ as compared to rule-based simplification if we use \texttt{Add} and \texttt{Mul} described in Section \ref{sec:generality}. Further details on this benchmark can be found in the repository\footnote{\url{https://github.com/JuliaSymbolics/SymbolicUtils.jl/pull/154#issuecomment-754302695}}.

\textbf{E-graph-based optimization for fewer CPU cycles} We developed a set of equality rules and a cost function to generate mathematically-equivalent expressions which are computed in fewer CPU cycles. To test the effectiveness of these techniques in a real-world scenario, we used the BioNetGen format to read in a 1122 ODE model of B Cell Antigen Signaling \cite{barua2012computational} and simplified the 24388 terms using Symbolics.jl. The generated code (explained in Section \ref{sec:codegen}) for the right-hand side accelerated from 15.023 $\mu$s to 7.372 $\mu$s per execution after the optimization, effectively halving the time required to solve the highly stiff ODEs. We have published the code for reproducing this benchmark in a Github Gist\footnote{\url{https://gist.github.com/shashi/a696020c6e65e1a3abfdbd74a3e6909c}}.

%

\textbf{Tracing and optimizing a PDE discretization.} 
When coupled with the ability to automatically generate symbolic expressions by tracing Julia code, this code generation process becomes an effective automated code optimization tool. We took a reaction-diffusion PDE discretization code written in native Julia, evaluated it with symbols, extracted the Jacobian in the form of a sparse matrix (rather than letting the solver use automatic-differentiation), generated codes containing multi-threaded parallelism, and achieved an overall 157x improvement in the solver time. This process required approximately 5 lines of Symbolics.jl code added to the PDE solver\footnote{ModelingToolkit.jl was created to automate the application of these optimizations to differential equation models~\cite{modelingtoolkit}}. The fully reproducible example can be found in the tutorial section of the documentation\footnote{Tutorial: ``Automated Sparse Parallelism of Julia Functions via Tracing'' (v0.1)}.

\section{Acknowledgements}

This material is based upon work supported by the National Science Foundation under grant no. OAC-1835443, grant no. SII-2029670, grant no. ECCS-2029670, grant no. OAC-2103804, and grant no. PHY-2021825. We also gratefully acknowledge the U.S. Agency for International Development through Penn State for grant no. S002283-USAID. The information, data, or work presented herein was funded in part by the Advanced Research Projects Agency-Energy (ARPA-E), U.S. Department of Energy, under Award Number DE-AR0001211 and DE-AR0001222. In part by DARPA under agreement number HR0011-20-9-0016 (PaPPa). We also gratefully acknowledge the U.S. Agency for International Development through Penn State for grant no. S002283-USAID. The views and opinions of authors expressed herein do not necessarily state or reflect those of the United States Government or any agency thereof. This material was supported by The Research Council of Norway and Equinor ASA through Research Council project "308817 - Digital wells for optimal production and drainage". Research was sponsored by the United States Air Force Research Laboratory and the United States Air Force Artificial Intelligence Accelerator and was accomplished under Cooperative Agreement Number FA8750-19-2-1000. The views and conclusions contained in this document are those of the authors and should not be interpreted as representing the official policies, either expressed or implied, of the United States Air Force or the U.S. Government. The U.S. Government is authorized to reproduce and distribute reprints for Government purposes notwithstanding any copyright notation herein.

\bibliographystyle{unsrt}
\bibliography{main}




\end{document}